\documentclass[conference]{IEEEtran}
\IEEEoverridecommandlockouts
\usepackage{cite}
\usepackage{amsmath,amssymb,amsfonts}
\usepackage{algorithmic}
\usepackage{graphicx}
\usepackage{textcomp}
\usepackage{xcolor}
\usepackage{multirow}
\usepackage{CJKutf8}

\usepackage[ruled,linesnumbered]{algorithm2e}

\usepackage[letterpaper,top=72pt,bottom=54pt,left=54pt,right=54pt]{geometry}

\def\BibTeX{{\rm B\kern-.05em{\sc i\kern-.025em b}\kern-.08em
    T\kern-.1667em\lower.7ex\hbox{E}\kern-.125emX}}
\begin{document}

\makeatletter
\newcommand{\linebreakand}{%
  \end{@IEEEauthorhalign}
  \hfill\mbox{}\par
  \mbox{}\hfill\begin{@IEEEauthorhalign}
}
\makeatother

\title{
FR-SLAM: A SLAM Improvement Method Based on Floor Plan Registration\\
\thanks{$^{*}$Xinde Li is corresponding author: {\tt\small xindeli@seu.edu.cn}}
}

\author{\IEEEauthorblockN{Jiantao Feng}
\IEEEauthorblockA{
\textit{Suzhou Campus} \\
\textit{Southeast University} \\
\textit{Nanjing center for applied mathematics}\\
Nanjing, China \\
fengjiantao1024@163.com}
\and
\IEEEauthorblockN{Xinde Li*}
\IEEEauthorblockA{
\textit{School of Automation} \\
\textit{School of Cyber Science and Engineering} \\
\textit{Shenzhen Research Institute} \\
\textit{Southeast University} \\
\textit{Nanjing center for applied mathematics}\\
Nanjing, China \\
xindeli@seu.edu.cn}
\and
\IEEEauthorblockN{HyunCheol Park}
\IEEEauthorblockA{\textit{Samsung Electronics} \\
South Korea \\
hc79.park@samsung.com}
\linebreakand
\IEEEauthorblockN{Juan Liu}
\IEEEauthorblockA{\textit{Samsung Electronics(China) R\&D Center} \\
Nanjing, China \\
juan82.liu@samsung.com}
\and
\IEEEauthorblockN{Zhentong Zhang}
\IEEEauthorblockA{
\textit{School of Cyber Science and Engineering} \\
\textit{Southeast University} \\
\textit{Nanjing center for applied mathematics}\\
Nanjing, China \\
zhentong\_zhang@seu.edu.cn}
}

\maketitle

\begin{abstract}

Simultaneous Localization and Mapping (SLAM) technology enables the construction of environmental maps and localization, serving as a key technique for indoor autonomous navigation of mobile robots. Traditional SLAM methods typically require exhaustive traversal of all rooms during indoor navigation to obtain a complete map, resulting in lengthy path planning times and prolonged time to reach target points. Moreover, cumulative errors during motion lead to inaccurate robot localization, impacting navigation efficiency. This paper proposes an improved SLAM method, FR-SLAM, based on floor plan registration, utilizing a morphology-based floor plan registration algorithm to align and transform original floor plans. This approach facilitates the rapid acquisition of comprehensive motion maps and efficient path planning, enabling swift navigation to target positions within a shorter timeframe. To enhance registration and robot motion localization accuracy, a real-time update strategy is employed, comparing the current position's building structure with the map and dynamically updating floor plan registration results for precise localization. Comparative tests conducted on real and simulated datasets demonstrate that, compared to other benchmark algorithms, this method achieves higher floor plan registration accuracy and shorter time consumption to reach target positions.
\end{abstract}

\begin{IEEEkeywords}
SLAM, Image registration, Floor plan, Morphological transformation
\end{IEEEkeywords}

\section{Introduction}

\begin{figure}[htbp]
    \begin{center}
    \includegraphics[width=3in]{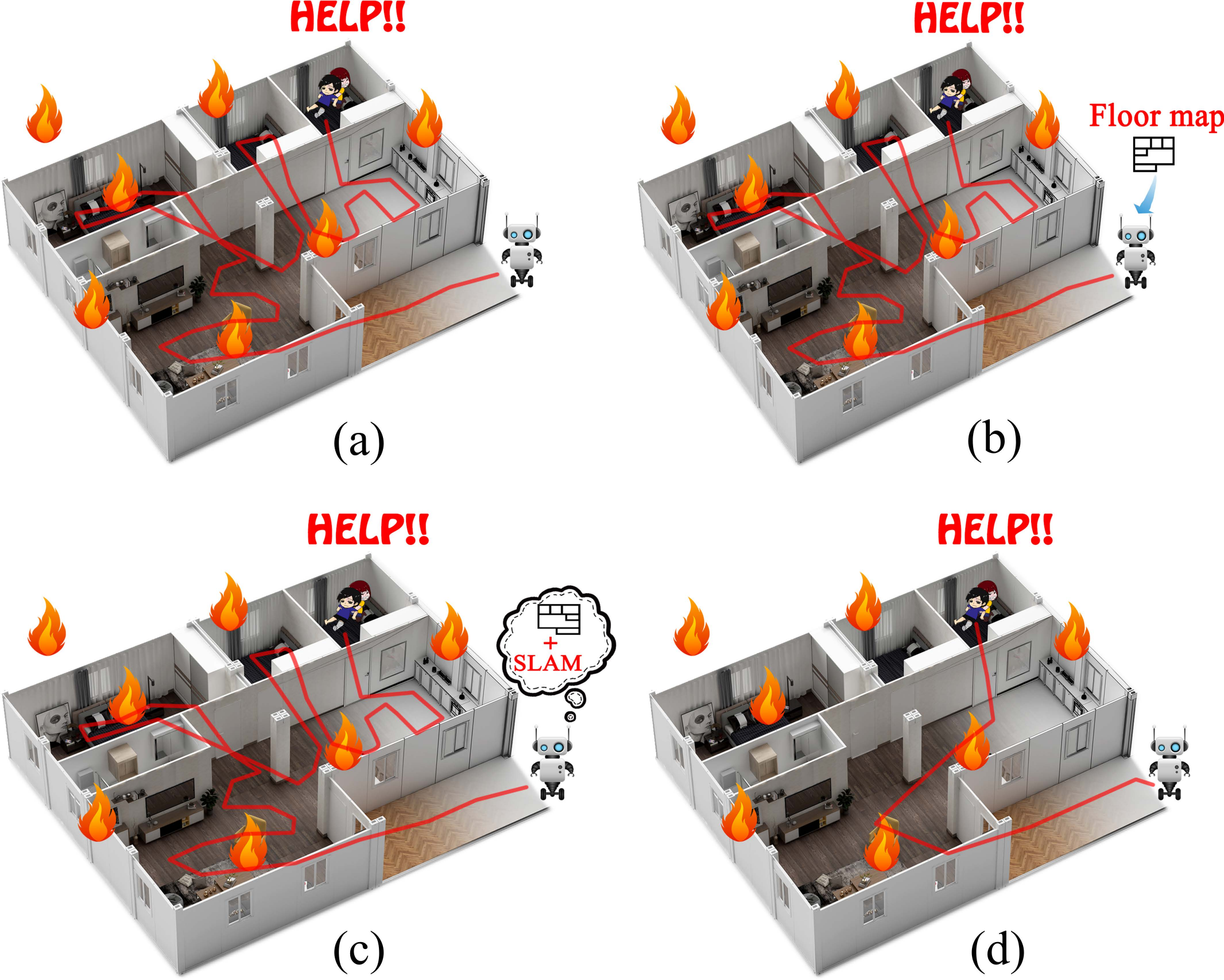}\\
    \caption{Schematic diagram of FR-SLAM algorithm}\label{FRSLAM_algorithm}
    \end{center}
    \vspace{-0.4cm}
\end{figure}

\begin{figure*}
    \begin{center}
    \includegraphics[width=7in]{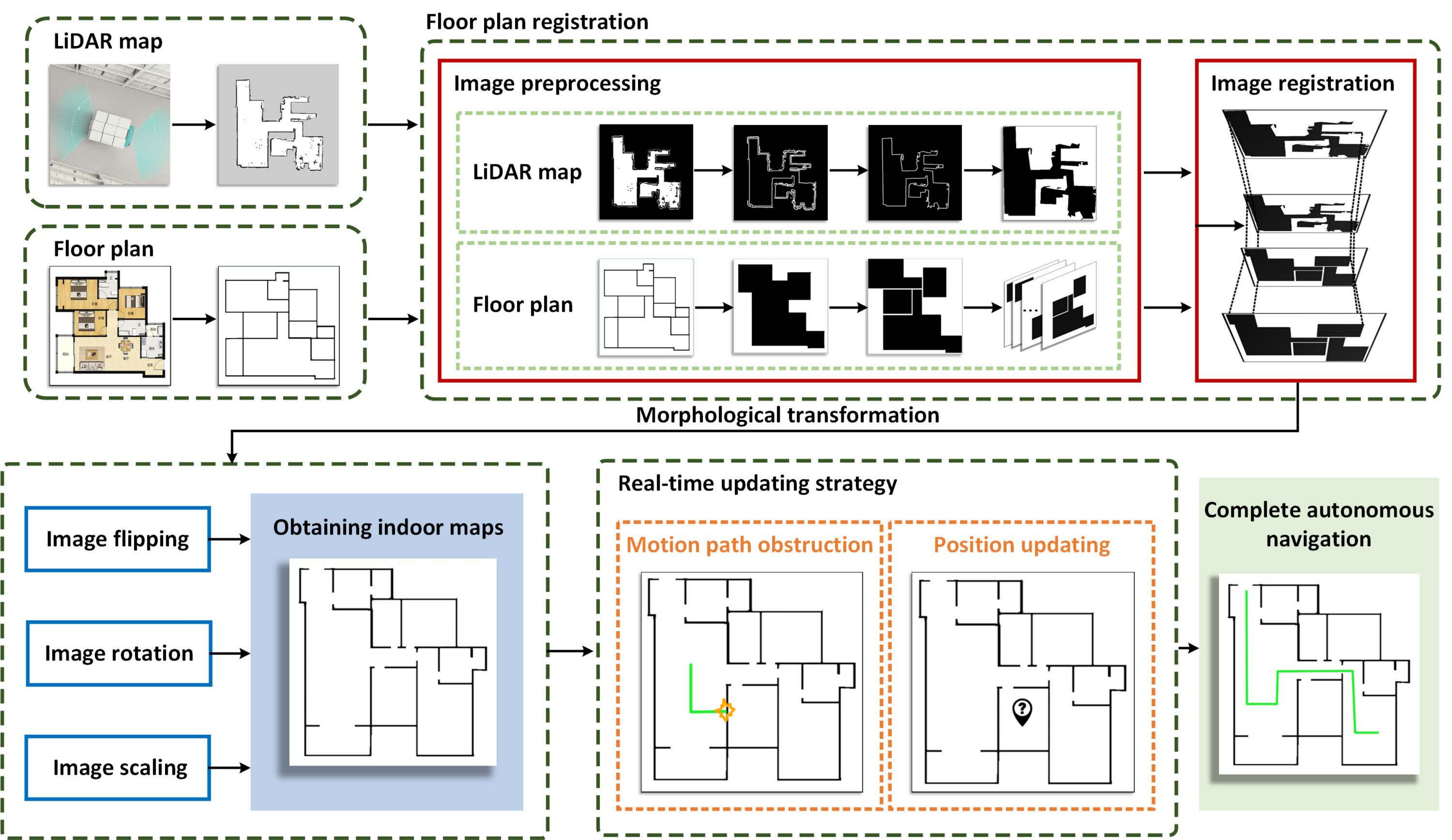}\\
    \caption{The overall workflow of FR-SLAM based on floor plans registration, including LiDAR map generation, original floor plans registration, and map update strategy.}\label{overall_workflow}
    \end{center}
    \vspace{-0.4cm}
\end{figure*}

In recent years, with the accelerated process of urbanization, the rapid expansion of urban areas, and the increasing accident rates in residential areas, the use of publicly available floor plans as references for search and rescue tasks by firefighters has become common. However, these floor plans often suffer from anomalies in aspect ratios, as well as issues related to rotation and flipping, which can adversely affect planning efforts. Concurrently, the advancement in miniaturization and intelligence of unmanned robotic systems has significantly contributed to tasks such as firefighting rescue operations \cite{b1} and post-disaster search and rescue missions. Equipped with laser rangefinders and utilizing ultrasonic sensors for high-precision ranging, indoor localization, and complex environmental perception, these robots play a vital role in map construction and path planning. Traditional SLAM methods typically involve scanning the entire indoor structure to complete global mapping, followed by loop closure detection to generate a complete map. However, obstacles such as closed doors and debris blocking pathways may prevent robots from accessing certain rooms, resulting in missing map information. These factors render conventional SLAM methods inadequate for real-time tasks such as firefighting rescue and post-disaster search and rescue operations.

This paper presents an improved SLAM method based on floor plan registration. The approach utilizes partial room LiDAR scans obtained by the robot as a reference, employing morphological transformations for image registration of the original floor plans (MFR). This enables the rapid acquisition of an authentic indoor map and facilitates path planning within a short timeframe, as depicted in Fig.~\ref{FRSLAM_algorithm}. Furthermore, it enhances the accuracy of robot motion localization through a real-time matching update strategy. The contribution of this work are as follows:

\begin{itemize}
\item This paper presents a SLAM improvement method based on floor plan registration, which integrates the original floor plan with LiDAR maps through image fusion to address the issues of accumulated errors during mapping and motion localization of the robot.

\item In this method, a registration algorithm for floor plans and LiDAR maps is proposed. The algorithm only requires contour information from images, and based on the maximum intersection over union (IoU) of floor plans and LiDAR maps under different morphological transformations, image registration can be accomplished, irrespective of environmental constraints. The accuracy of registration achieved by this algorithm on the testing dataset reaches 85$\%$, surpassing common image matching algorithms.

\item To address the problem of robot motion localization, a real-time matching update strategy is proposed. This strategy continuously matches the current LiDAR map with the floor plan, eliminating motion errors and achieving real-time localization.
\end{itemize}

\section{RELATED WORK}
Image registration is widely used in various fields such as robot visual navigation \cite{b3}, motion planning \cite{b4}, medical image processing \cite{b5}, and remote sensing. However, due to the diverse types of images applicable for image registration, there is currently no universal optimization method that can satisfy all purposes \cite{b6}. Commonly used image registration methods include intensity-based matching algorithms and feature-based matching algorithms. Intensity-based image registration methods are sensitive to changes in image intensity and require images to have rich intensity information. On the other hand, feature-based matching algorithms require the extraction of image features such as closed area areas, image contours, key intersections, and corners, which necessitates images with abundant and easily extractable relevant features.

The SLAM method encompasses two primary tasks: localization and mapping. A robot begins from an unknown location in an unknown environment and uses repeated observations of environmental features during movement to determine its position and orientation. Subsequently, it incrementally constructs a map of the surrounding environment based on its location, thereby achieving the objectives of simultaneous localization and mapping \cite{Alsadik2021}. Traditional indoor SLAM methods require the robot to perform a complete scan of the entire scene, which is time-consuming and yields poor localization accuracy. 


\section{METHOD}

This section presents the complete workflow of FR-SLAM (illustrated in  Fig.~\ref{overall_workflow}), which includes: image registration algorithm comprising image preprocessing and registration, aligning the original floor plan with LiDAR map, applying morphological transformations to the floor plan based on registration results, thereby obtaining a floor plan that describes the structure of the house as the robot's motion map; real-time update strategy, involving updating LiDAR information to refine the alignment of the floor plan with the actual house structure, mitigating obstacles to robot motion. Additionally, it compares the current location's house structure with the motion map to enhance the accuracy of robot motion localization.

\subsection{Floor Plan Registration}\label{section_floor_plan_registration}
\subsubsection{Image Preprocessing}
Image preprocessing consists of two parts: floor plan preprocessing and LiDAR map preprocessing.

The purpose of floor plan preprocessing is to generate multiple filled images covering different room arrangements, including living rooms and various other rooms, to address the missing areas of different rooms in the LiDAR map. The formula is as follows:
\begin{equation}\label{eq.area}
\begin{array}{ll}
    F=\{f(S): S \subseteq H\}
\end{array}
\end{equation}
where $\mathit{H}$ represents all room types and $\mathit{F}$ represents the final filled image. The specific steps are shown in Alg.~\ref{floor_preprocessing}.

\begin{algorithm}[h]
\caption{Floor plan preprocessing algorithm}\label{floor_preprocessing}
\SetAlgoLined
\KwIn{Original floor plan room list $room\_list$}
\KwOut{Floor plan collection $floor\_list$}
\BlankLine
Initialize bedroom dictionary $bedroom\_map = \text{\{ \}}$\;
Initialize $temp\_list = \text{[ ]}$\;
\For{$i=0$ \KwTo $length(room\_list)$}{
    Add bedroom to $bedroom\_map$\;
}
\For{$i=0$ \KwTo $length(bedroom\_map.keys())$}{
    Calculate different bedroom arrangement combinations $combin\_list$\;
    \For{$j=0$ \KwTo $length(combin\_list)$}{
        $temp\_list$.add($combin\_list[j] + living\_room$)\;
    }
}
\For{$i=0$ \KwTo $length(temp\_list)$}{
    $temp\_floor \leftarrow$ draw floor plan($temp\_list(i)$)\;
    $floor\_list$.add($temp\_floor$)\;
}
\KwRet{$floor\_list$}\;
\end{algorithm}

The preprocessing of LiDAR images involves initially converting the original grayscale image $\mathit{I}$ into a binary image $\mathit{B}$. The binarization process can be represented by the following equation:
\begin{equation}\label{eq.area}
\begin{array}{ll}
    B(x, y)= \begin{cases}1 & I(x, y) \geqslant \theta \\ 0 & I(x, y) \leqslant \theta\end{cases}
\end{array}
\end{equation}
where $\mathit{\theta}$ represents the grayscale threshold value of the pixel. Subsequently, connected component analysis is performed on the binary image $\mathit{B}$ to remove contours with small areas, retaining the main structure of the LiDAR image, represented using set notation as:
\begin{equation}\label{eq.area}
\begin{array}{ll}
    B^{\prime}=\{p \in B: \operatorname{area}(p) \geq \alpha\}
\end{array}
\end{equation}
where $\mathit{\alpha}$ represents the contour threshold, and 
$\mathit{area(p)}$ denotes the area of connected domain $\mathit{p}$.

Subsequently, polygon fitting is employed to smooth the outer contours, thereby reducing computational complexity. Polygon fitting typically utilizes the least squares method. Given a set containing $\mathit{area(n)}$ points ${\text{[}(x_1, y_1), (x_2, y_2), …, (x_n, y_n)\text{]}}$, the fitting problem is transformed into solving a system of linear equations:
\begin{equation}\label{eq.area}
\begin{array}{ll}
    \left[\begin{array}{ccccc}
1 & x_1 & x_1^2 & \cdots & x_1^m \\
1 & x_2 & x_2^2 & \cdots & x_2^m \\
\vdots & \vdots & \vdots & \ddots & \vdots \\
1 & x_n & x_n^2 & \cdots & x_n^m
\end{array}\right]\left[\begin{array}{c}
a_0 \\
a_1 \\
\vdots \\
a_m
\end{array}\right]=\left[\begin{array}{c}
y_1 \\
y_2 \\
\vdots \\
y_n
\end{array}\right]
\end{array}
\end{equation}
By solving this system, polynomial coefficients are obtained, which are then used to derive the fitted polygon $\mathit{f(x)}$. Finally, the flood fill algorithm is applied to fill the surrounding area, and the image is cropped, completing the preprocessing of the LiDAR image.

\subsubsection{Image Registration}


The preprocessed floor plan and LiDAR map are fused and registered in the image, as depicted in Alg.
~\ref{floor_registration}, following a comprehensive process.

\begin{algorithm}[h]
\caption{Floor plan registration algorithm}\label{floor_registration}
\KwIn{LiDAR image $L$; Original floor plan image $F$}
\KwOut{Rotation result $R_r$, Flip result $R_f$, Horizontal scaling result $R_h$, Vertical scaling result $R_v$}
$L^* \leftarrow \text{resize}(L, 200, 200)$\;
$F^* \leftarrow \text{resize}(F, 200, 200)$\;
$list\_F^* \leftarrow$ Morphological transformations on floor plan image $F^*$\;
Initialize $list\_IoU = \text{[ ]}$\;
\For{$i=0$ \KwTo $\text{length}(list\_F^*)$}{
    $S_\text{inter}, S_\text{union} \leftarrow$ Fuse $R^*$ and $list\_F^*\text{[i]}$, calculate intersection area and union area according to Formula (5)\;
    $temp\_IoU \leftarrow$ Calculate IoU($S_\text{inter}$, $S_\text{union}$) according to Formula (6)\;
    $list\_IoU.\text{add}(temp\_IoU)$\;
}
$R_r, R_f \leftarrow$ Transformation corresponding to maximum IoU ($list\_IoU$)\;
$F^{**} \leftarrow$ Deform $F$ according to transformation parameters ($R_r, R_f$)\;
$R_h, R_v \leftarrow$ Calculate horizontal and vertical scaling ratios according to Formula (7) ($L, F^{**}$)\;
\KwRet{$R_r, R_f, R_h, R_v$}\;
\end{algorithm}

When computing the intersection and union image areas, the Harris operator is employed to locate the contour vertices of the image, resulting in a vertex set of polygons $A = {\text{[}(x_1, y_1), (x_2, y_2), ..., (x_n, y_n)\text{]}}$. The Gaussian area formula is utilized to calculate the area of this irregular polygon:
\begin{equation}\label{eq.area}
\begin{array}{ll}
    S=\frac{1}{2}\left|\sum_{i=1}^n\left(x_i y_{i+1}-x_{i+1} y_i\right)\right|
\end{array}
\end{equation}
After obtaining the intersection area $S_{inter}$ and union area $S_{union}$ of the images, the IoU is calculated as follows:
\begin{equation}\label{eq.iou}
\begin{array}{ll}
    \text { IoU }=S_{\text {inter }} / S_{\text {union }}
\end{array}
\end{equation}
Once the maximum IoU is found, the transformation parameters for folding and rotating the floor plan can be determined. Morphological transformations are applied to the corresponding combinations of floor plans to obtain the maximum width and height $(H_{1}$, $V_{1})$ of the floor plan and the maximum width and height $(H_{2}$, $V_{2})$ of the original LiDAR map. The horizontal and vertical scaling ratios are then calculated as follows:
\begin{equation}\label{eq.area}
\begin{array}{ll}
    \left\{\begin{array}{l}
S_{\mathrm{H}}=H_1 / H_2 \\
S_{\mathrm{V}}=V_1 / V_2
\end{array}\right.
\end{array}
\end{equation}
where $S_{H}$ represents the horizontal scaling ratio and $S_{V}$ represents the vertical scaling ratio.This completes the registration of the original floor plan with reference to the LiDAR map.

\subsection{Real-time updating strategy}
Upon completing the registration of the original floor plan, the real floor plan can be obtained by applying image transformations based on the registration parameters. This transformed result serves as the robot's motion map, utilized for executing path planning tasks.

However, in dealing with complex room structures, there may be instances where several rooms share similar configurations. In such cases, the indoor map obtained through floor plan registration may be erroneous. To address this issue, it is necessary to employ a new LiDAR map covering a larger area of the rooms and conduct floor plan registration again to update the indoor map. Additionally, after the robot has traveled a certain distance, it needs to reposition itself and update its location on the map to ensure route alignment. To tackle these challenges, this paper introduces a real-time map updating strategy, as outlined in Alg.~\ref{map_updating}

\begin{algorithm}[h]
    \caption{Real-time Map Updating Algorithm}\label{map_updating}
    \SetAlgoLined
    \KwIn{LiDAR map $L$; Original floor plan $F$; Relocation distance $D$}
    \KwOut{Motion map $M$; Current robot position $P$}
    Initialize robot travel distance $d=0$\;
    Initialize remaining travel distance $S$\;
    Initialize motion map $M$\;
    Initialize partial room structure $N \leftarrow$ Current position partial room structure\;
    Initialize robot position $P$\;
    \If{$M$ is not available}{
        $M \leftarrow$ Floor plan registration as described in Section~\ref{section_floor_plan_registration} ($L, F$)\;
    }
    $P \leftarrow$ Obtain current robot position ($N, M$)\;
    $S \leftarrow$ Path planning, calculate remaining travel distance ($P, M$)\;
    \While{$S > 0$}{
        $N \leftarrow$ Current position partial room structure\;
        $L \leftarrow$ LiDAR map updating ($L, N$)\;
        // When robot motion is obstructed or exceeds relocation distance\;
        \If{$(d \leq 0$ or $d \geq D)$}{
            $M \leftarrow$ Floor plan registration ($L, F$)\;
            $P \leftarrow$ Obtain current robot position ($N, M$)\;
            $S \leftarrow$ Path planning, calculate remaining travel distance ($P, M$)\;
            $d \leftarrow 0$\;
        }
        $d \leftarrow d +$ robot's previous second travel distance\;
        $S \leftarrow S -$ robot's previous second travel distance\;
    }
    \Return $M, P$\;
\end{algorithm}

Upon obtaining the indoor map with the floor plan as reference, the robot commences movement along the planned path. During motion, the robot continues to scan the surrounding environment and update the LiDAR map. If the motion path is obstructed or after traveling a certain distance, the current position $x_t$ is estimated based on the motion model:
\begin{equation}\label{eq.area}
\begin{array}{ll}
     x_t=g\left(u_t, x_{t-1}\right)+\varepsilon_t
\end{array}
\end{equation}
where $g$ is the robot’s motion model function, $u_t$ is the control input at time $t$, $x_{t-1}$ is the robot’s position at the previous moment, $\varepsilon_t$ represents the motion noise. Simultaneously, the indoor map is updated by registering the current LiDAR map with the floor plan. Finally, based on the current room structure, the corresponding position coordinates are identified on the map to achieve precise localization, thereby allowing for path planning again. This workflow iterates until the task is completed.

\section{EXPERIMENTS}
\subsection{Dataset Construction}
The LiDAR images within the dataset are derived from two sources: some are generated via simulations in a GazeBo environment, while others are captured by robots operating in real-world settings, as illustrated in Fig.~\ref{LiDAR_map_obtained}.

\begin{figure}[htbp]
    \begin{center}
    \includegraphics[width=3in]{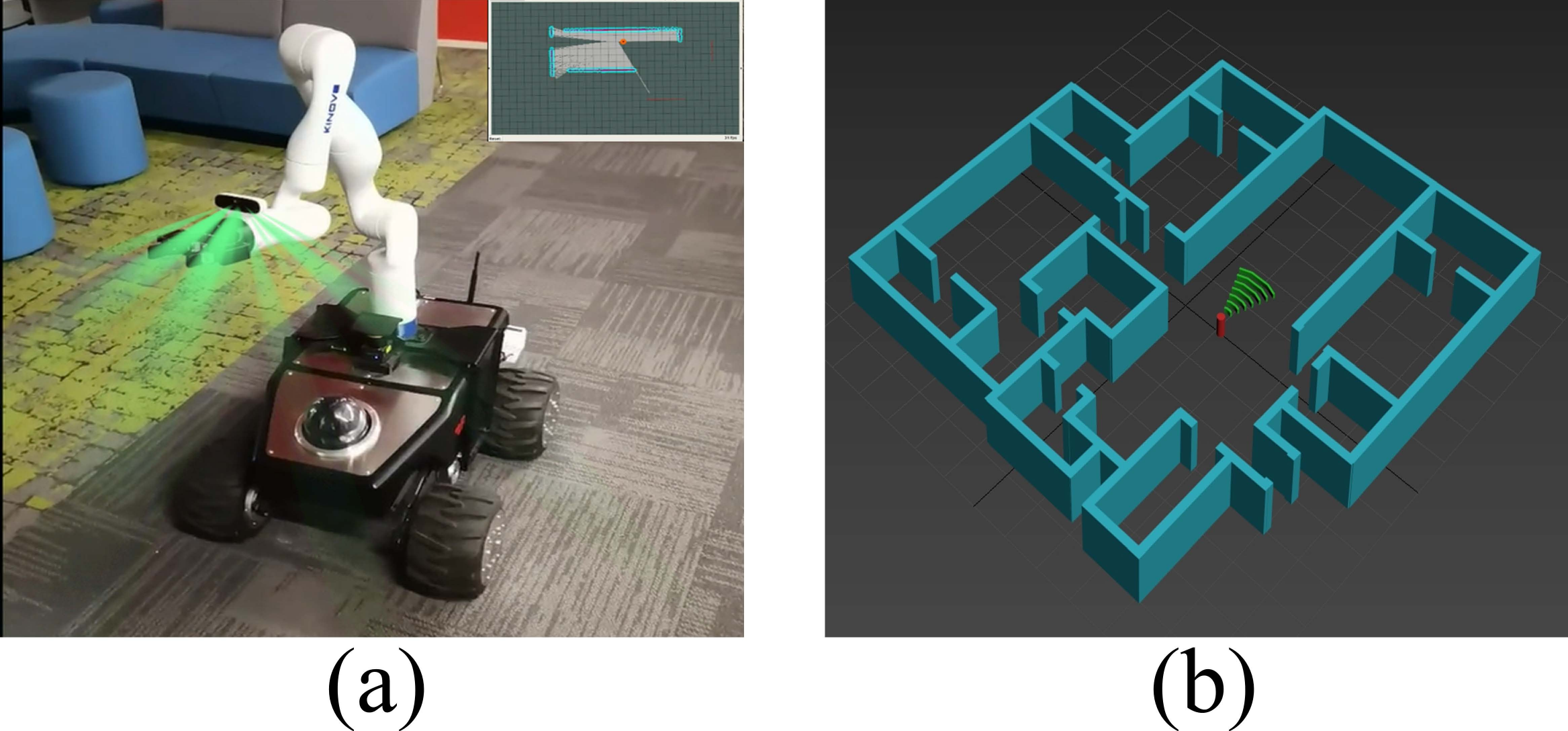}\\
    \caption{(a) The LiDAR map obtained by the robot running in the actual environment (b) Simulation environment generating LiDAR map}\label{LiDAR_map_obtained}
    \end{center}
    \label{f04}
    \vspace{-0.2cm}
\end{figure}

The floor plans in the dataset also originate from two distinct sources: a portion is collected from commonly available floor plans on the internet, and another portion consists of manually drawn representations of actual room layouts. All floor plans are stored in JSON format, detailing the coordinates of corner points of each room, as depicted in Fig.~\ref{floor_plan_obtained}.

\begin{figure}[htbp]
    \begin{center}
    \includegraphics[width=3in]{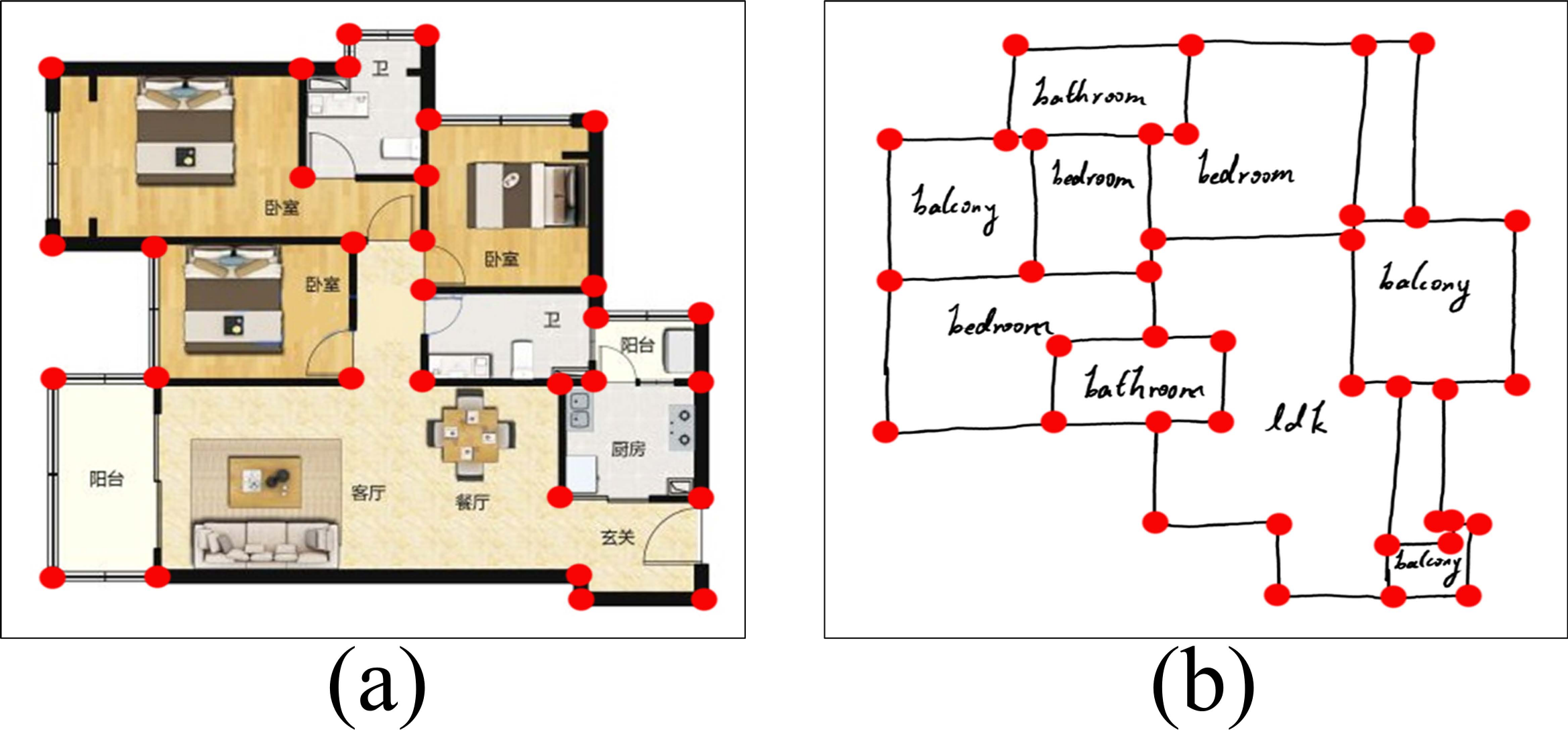}\\
    \caption{(a) Annotate the floor plan obtained from the network (b) Annotate the manually drawn floor plan}\label{floor_plan_obtained}
    \end{center}
    \vspace{-0.4cm}
\end{figure}

\subsection{Comparison of registration}
This paper employs the average IoU ($IoU_a$) of the morphological transformed floor plans as a crucial metric for evaluating the accuracy of the registration algorithm. The IoU of each room is calculated according to (6), followed by the computation of $IoU_a$:
\begin{equation}\label{eq.area}
\begin{array}{ll}
     IoU_a=SUM_{IoU} / n
\end{array}
\end{equation}
where $SUM_{IoU}$ is the sum of all room IoU, $n$ is the number of rooms.



\begin{figure}[htbp]
  \begin{center}
  \includegraphics[width=3.2in]{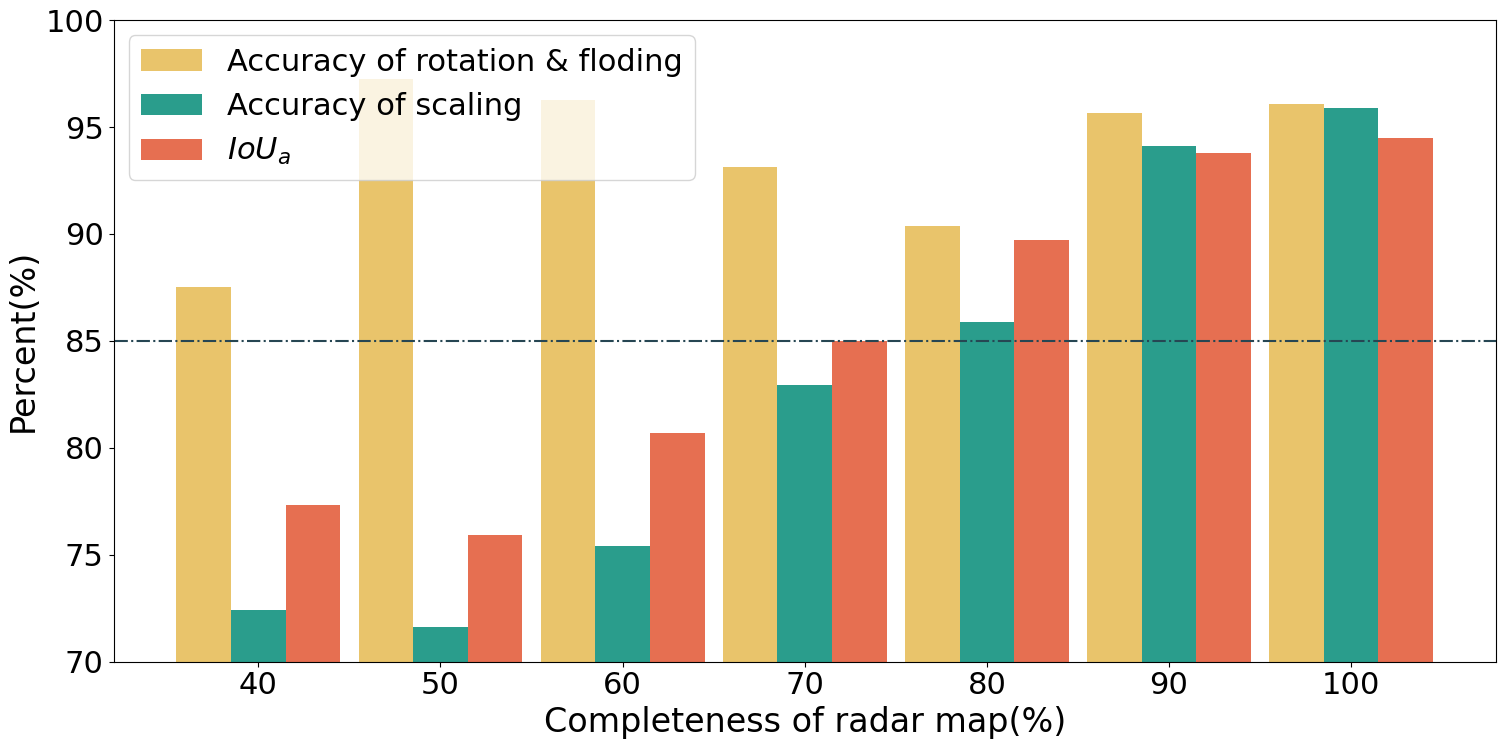}\\
  \caption{The folding accuracy, rotation accuracy, and $IoU_a$ are evaluated for different degrees of completeness in the LiDAR map}  \label{compar_registration_fig}
  \end{center}
\end{figure}

The performance of the proposed algorithm on the test dataset is depicted in Fig.~\ref{compar_registration_fig}. Notably, under various degrees of completeness in LiDAR images, the accuracy of rotation and flipping consistently exceeds 85$\%$. The primary factor influencing the $IoU_a$ is the accuracy of scaling, exhibiting a similar trend to that of scaling accuracy. The $IoU_a$ of the registered floor plans gradually increases with the completeness of the LiDAR images, ultimately surpassing 94$\%$, while the overall $IoU_a$ exceeds 85$\%$, demonstrating substantial practical utility of the algorithm.

To compare the effectiveness of different image registration algorithms, experiments were conducted using SURF \cite{b7}, OBR \cite{b8}, NBB \cite{b9}, DIRNet \cite{b10}, and the MFR algorithm. The algorithms were applied for image registration, and the runtime of each algorithm was recorded.

\begin{table}[]
\caption{Comparing the performance of different matching algorithms}\label{table_performance_of_different_matching_algorithms}
\label{t_compare}
\setlength{\tabcolsep}{2.2mm}
\begin{tabular}{ccccc}
\hline
Algorithms   & \begin{tabular}[c]{@{}c@{}}Fold\\ Accuracy(\%)\end{tabular} & \begin{tabular}[c]{@{}c@{}}Rotation\\ Accuracy(\%)\end{tabular} & $IoU_a$(\%)         & \begin{tabular}[c]{@{}c@{}}Average\\ Time(s)\end{tabular} \\ \hline
SURF         & 45.97                                                       & 43.57                                                           & 43.49          & 0.45                                                   \\
ORB          & 53.75                                                       & 59.65                                                           & 53.18          & \textbf{0.38}                                          \\
NBB          & 75.98                                                       & 77.36                                                           & 74.85          & 1.73                                                   \\
DIRNet       & 68.37                                                       & 66.38                                                           & 63.76          & 1.52                                                   \\
\textbf{MFR} & \textbf{87.71}                                              & \textbf{96.71}                                                  & \textbf{85.12} & 0.92                                                   \\ \hline
\end{tabular}
\vspace{-0.4cm}
\end{table}

The results, as shown in Table~\ref{table_performance_of_different_matching_algorithms}, indicate that the registration algorithm proposed in this paper outperforms other commonly used image matching algorithms in terms of folding accuracy, rotation accuracy, and $IoU_a$ on the test dataset. However, this algorithm consumes a relatively longer time, with an average runtime of 0.92 seconds per match, higher than traditional image registration algorithms but lower than deep learning algorithms. Considering the general requirement of real-world applications for runtime to be within 1 second, this algorithm meets the requirement. Through comprehensive analysis, it is concluded that this algorithm has significant advantages in solving the image registration problem between the original floor plan and LiDAR map.

\subsection{Simulated Rescue Scenario Testing}
Using GazeBo, multiple indoor simulated search and rescue scenarios were constructed based on the HM3D dataset \cite{b11} to comprehensively evaluate the effectiveness of the proposed improvement method. In this study, comparisons were made with E2DG-SLAM \cite{b12}, UV-SLAM \cite{b13}, and TLIO-SLAM \cite{b14}. The simulated rescue scenarios encompass three situations: the first involves a single target indoors with a known orientation; the second involves multiple targets indoors with known orientations; and the third involves an unknown number of targets indoors with unknown orientations.

\begin{figure}[htbp]
  \begin{center}
  \includegraphics[width=3.4in]{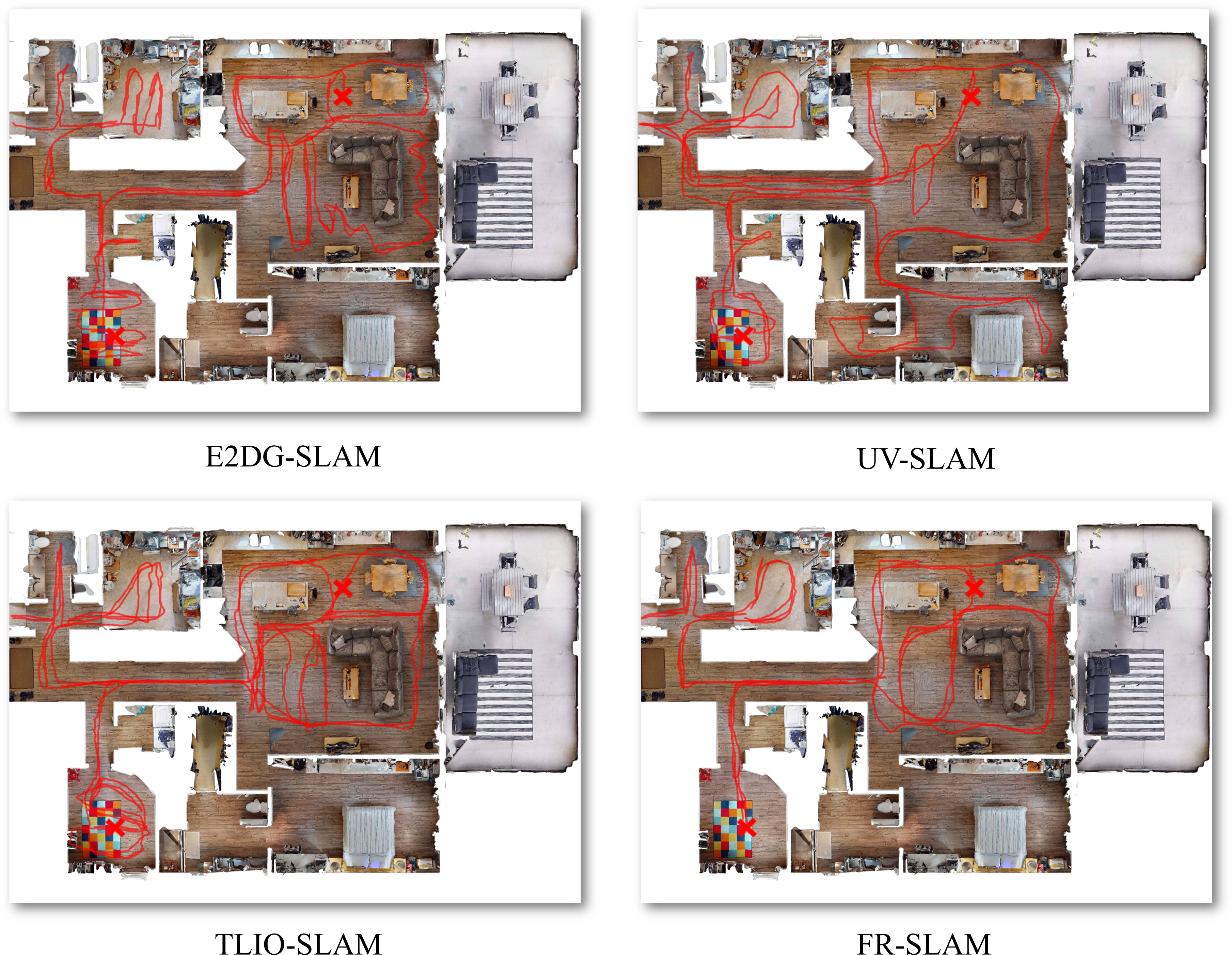}\\
  \caption{Motion trajectory maps of different SLAM algorithms}\label{motion_trajectory_maps}
  \end{center}
\end{figure}

Fig.~\ref{motion_trajectory_maps} illustrates the trajectory results of different SLAM algorithms in finding two targets with known orientations. It can be intuitively observed that the FR-SLAM algorithm, after completing the registration of the floor plan and obtaining the overall room map, can directly reach the position of the second target. Table~\ref{table_time_consumption_slam} presents the average rescue times of different SLAM algorithms in three simulated rescue scenarios. It can be seen that the FR-SLAM algorithm consumes less time in different simulated scenarios compared to the other algorithms, especially when the number and orientation of targets are known, where it is 6.7$\%$ lower than the second-lowest method. This is because the other three algorithms, during the rescue task, do not know if the current room is connected to other rooms, and the location of the door to enter the next room is also unknown, thus requiring exploration of each room. The advance knowledge of the target orientation only changes the order of house search but does not allow skipping rooms. The search can only be halted when the number of found targets equals the number of targets known in advance.

\begin{table}[]
	\caption{The time consumption of different SLAM algorithms}
	\label{table_time_consumption_slam}
	\setlength{\tabcolsep}{3.8mm}
	\begin{tabular}{cccc}
		\hline
		\multirow{2}{*}{Algorithms} & \multicolumn{2}{c}{Direction known} & \multirow{2}{*}{Direction unknown} \\
		& Single           & Multiple         &                      \\ \hline
		E2DG-SLAM                   & 16m35s           & 28m49s           & 36m47s               \\
		UV-SLAM                     & 18m45s           & 27m15s           & 39m27s               \\
		TLIO-SLAM                   & 17m15s           & 29m38s           & 38m19s               \\
		\textbf{FR-SLAM}            & \textbf{16m12s}  & \textbf{25m24s}  & \textbf{34m53s}      \\ \hline
	\end{tabular}
	\vspace{-0.4cm}
\end{table}

\section{CONCLUSIONS}
This paper proposes an improved SLAM method, FRSLAM, based on floor plan registration, addressing the issues of prolonged mapping duration and inaccurate localization. Utilizing LiDAR images for aligning and deforming floor plans, a comprehensive motion map is obtained. Concurrently, a real-time update strategy is employed to enhance the accuracy of floor plan registration and robot motion localization in complex environments. The MFR algorithm proposed within the method achieve $IoU_a$ of 85$\%$ in comparative tests, surpassing other benchmark algorithms. In simulated rescue scenario tests, the FR-SLAM algorithm presented in this paper achieves a maximum time savings of 6.7$\%$ compared to other benchmark algorithms in completing rescue tasks. However, this algorithm is limited to two-dimensional floor plan information and does not incorporate the three-dimensional information generated by LiDAR point clouds, resulting in limited available information for image registration and robot localization. Future work aims to leverage point cloud information for three-dimensional reconstruction and utilize three-dimensional models for robot navigation to improve algorithm efficiency.

\end{document}